\documentclass[sigconf]{acmart}
\AtBeginDocument{%
  }

\setcopyright{acmlicensed}
\copyrightyear{2026}
\acmYear{2026}
\acmDOI{XXXXXXX.XXXXXXX}
\acmConference[Conference acronym 'XX]{Make sure to enter the correct
  conference title from your rights confirmation email}{June 03--05,
  2026}{Woodstock, NY}
\acmISBN{978-1-4503-XXXX-X/2026/06}

\usepackage{subcaption}
\usepackage{booktabs}
\usepackage{multirow}
\usepackage[table]{xcolor}
\usepackage{graphicx}
\usepackage{tabularx}
\usepackage{array}
\usepackage{caption}

\begin{document}

\title{One Model, Two Minds: Task-Conditioned Reasoning for Unified Image Quality and Aesthetic Assessment}
\author{Wen Yin}
\orcid{0009-0009-6099-9091}
\affiliation{%
  \institution{University of Electronic Science and Technology of China}
  \city{Chengdu}
  \state{Sichuan}
  \country{China}
}
\affiliation{%
  \institution{Jiigan Technology}
  \city{Chengdu}
  \state{Sichuan}
  \country{China}}
\email{yinwen1999@std.uestc.edu.com}

\author{Cencen Liu}
\affiliation{%
  \institution{University of Electronic Science and Technology of China}
  \city{Chengdu}
  \state{Sichuan}
  \country{China}
}
\affiliation{%
  \institution{Jiigan Technology}
  \city{Chengdu}
  \state{Sichuan}
  \country{China}}
\email{202411900402@std.uestc.edu.com}

\author{Dingrui Liu}
\author{Bing Su}
\affiliation{%
  \institution{Jiigan Technology}
  \city{Chengdu}
  \state{Sichuan}
  \country{China}}
\email{liudingrui@jiigan.com}
\email{subing@jiigan.com}

\author{Yuan-Fang Li}
\affiliation{%
  \institution{Faculty of Information Technology, Monash University}
  \city{Melbourne}
  \country{Australia}}
\email{yuanfang.li@monash.edu}

\author{Tao He}
\authornote{Corresponding author.}
\affiliation{%
  \institution{The Laboratory of Intelligent Collaborative Computing of UESTC}
  \city{Chengdu}
  \state{Sichuan}
  \country{China}}
\email{tao.he01@hotmail.com}

\renewcommand{\shortauthors}{Trovato et al.}


\begin{abstract}
Unifying Image Quality Assessment (IQA) and Image Aesthetic Assessment 
(IAA) in a single multimodal large language model is appealing, yet 
existing methods adopt a task-agnostic recipe that applies the same 
reasoning strategy and reward to both tasks. We show this is 
fundamentally misaligned: IQA relies on low-level, objective 
perceptual cues and benefits from concise distortion-focused 
reasoning, whereas IAA requires deliberative semantic judgment and is 
poorly served by point-wise score regression. We identify these as a 
\emph{reasoning mismatch} and an \emph{optimization mismatch}, and 
provide empirical evidence for both through controlled probes. 
Motivated by these findings, we propose TATAR (Task-Aware Thinking 
with Asymmetric Rewards), a unified framework that shares the 
visual-language backbone while conditioning post-training on each 
task's nature. TATAR combines three components: fast--slow 
task-specific reasoning construction that pairs IQA with concise 
perceptual rationales and IAA with deliberative aesthetic narratives; 
two-stage SFT+GRPO learning that establishes task-aware behavioral 
priors before reward-driven refinement; and asymmetric rewards that 
apply Gaussian score shaping for IQA and Thurstone-style completion 
ranking for IAA. Extensive experiments across eight benchmarks 
demonstrate that TATAR consistently outperforms prior unified 
baselines on both tasks under in-domain and cross-domain settings, 
remains competitive with task-specific specialized models, and yields 
more stable training dynamics for aesthetic assessment. Our results 
establish task-conditioned post-training as a principled paradigm for 
unified perceptual scoring. Our code is publicly available at \url{https://github.com/yinwen2019/TATAR}.
\end{abstract}

\maketitle
\section{Introduction}\label{sec:intro}

Images are judged on two distinct axes in the real world: 
\emph{technical quality} (whether blur, noise, or compression 
has degraded perceptual fidelity) and \emph{aesthetic appeal} 
(whether composition, lighting, and semantic expressiveness 
combine into something visually compelling). Image Quality 
Assessment (IQA)~\cite{RALI,VQ_R1,DeQA_Doc,DeQA_score} and 
Image Aesthetic Assessment 
(IAA)~\cite{photographer,AMHP,SRIAA,Humanaesexpert} formalize 
these two axes, and both are indispensable in modern visual 
pipelines spanning image acquisition~\cite{SPAQ}, 
transmission~\cite{transmission1,transmission2}, content 
recommendation~\cite{IAAsurvey}, and creative 
assistance~\cite{compositionIAA}. In practice, users routinely 
expect a single system to answer both: \emph{is this image 
technically good?} and \emph{is it beautiful?} This demand has 
motivated a growing line of work on unified IQA--IAA modeling 
with Multimodal Large Language Models 
(MLLMs)~\cite{Qwen25vl,Qwen3vl,llava15,Internvl3}.

\begin{figure}[t]
    \centering
    \includegraphics[width=0.95\columnwidth]{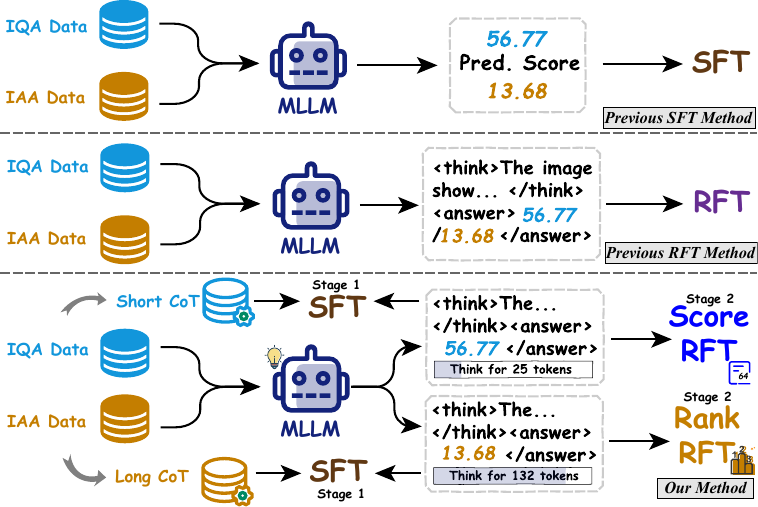}
    \caption{%
        \textbf{From shared to task-conditioned post-training.}
        Prior unified methods apply the same response behavior 
        and reward to both IQA and IAA, conflating two tasks 
        with fundamentally different decision regimes. TATAR 
        decouples \emph{what is shared} (the VL 
        backbone) from \emph{what is adapted} (post-training): 
        Stage~1 instills task-specific reasoning modes via 
        fast--slow SFT, and Stage~2 refines scoring with 
        asymmetric rewards matched to each task's supervision 
        geometry.%
    }
    \label{fig:model_comparison}
\end{figure}

Existing unified methods~\cite{Nexttoken,Qalign,UniPercept} follow a simple recipe: they share the same backbone and optimize both IQA and IAA with the same objective, either score regression~\cite{Qalign} or a single shared reinforcement reward~\cite{UniPercept}. This design is convenient, but it assumes the two tasks can be solved in the same way. In practice, they require different evidence and different aggregation. IQA is largely determined by low-level, relatively objective cues; accurate scoring often reduces to detecting distortions and mapping them to fidelity degradation~\cite{FRIQA,EvoQuality,MPD}. In contrast, IAA is preference-sensitive and depends more on high-level semantics, composition, lighting, and overall visual impression~\cite{RPCD,Charm}. Training them with a shared thinking style and a shared reward therefore introduces a mismatch: it either encourages unnecessary deliberation for IQA or provides an overly rigid regression signal for IAA, both of which can hinder unified learning.

Building on this, a key question is raised: \emph{in a unified IQA–IAA model, what should be shared and what should remain task-specific?} We highlight two concrete mismatches that prior unified training pipelines~\cite{UniPercept,Qalign} do not explicitly address:
\textbf{(i) Reasoning mismatch.} IQA typically benefits from concise distortion-focused judgment: once major artifacts are identified, longer reasoning rarely improves the score and can even distract from low-level evidence. IAA, in contrast, often benefits from more deliberation to integrate multiple aesthetic factors (e.g., composition, lighting, semantics) before producing a holistic rating. A single, uniform CoT style therefore tends to be suboptimal for at least one task.
\textbf{(ii) Optimization mismatch.} IQA annotations are designed to approximate an absolute fidelity scale, making point-wise score regression a natural objective. IAA annotations, however, capture subjective preference and exhibit higher ambiguity; treating them as equally precise regression targets can yield unstable optimization and poor alignment with relative human judgments.

\begin{figure}[t]
    \centering
    \begin{subfigure}[b]{0.49\linewidth}
        \centering
        \includegraphics[width=\linewidth]{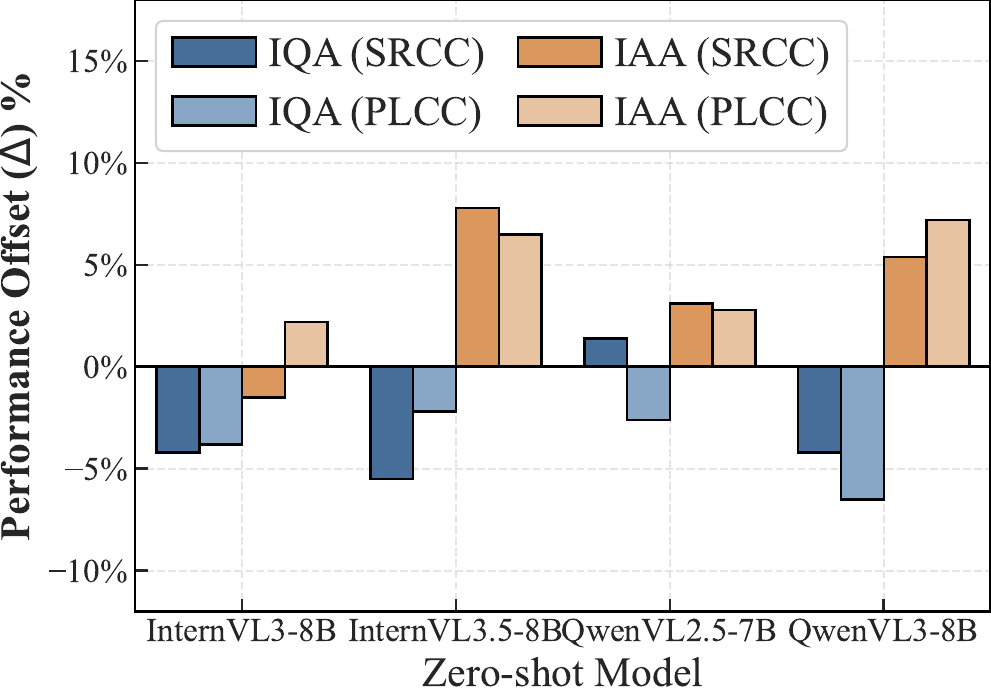}
        \caption{Task-dependent effect of CoT}
        \label{fig:CoTPerformanceOffset}
    \end{subfigure}
    \hfill
    \begin{subfigure}[b]{0.49\linewidth}
        \centering
        \includegraphics[width=\linewidth]{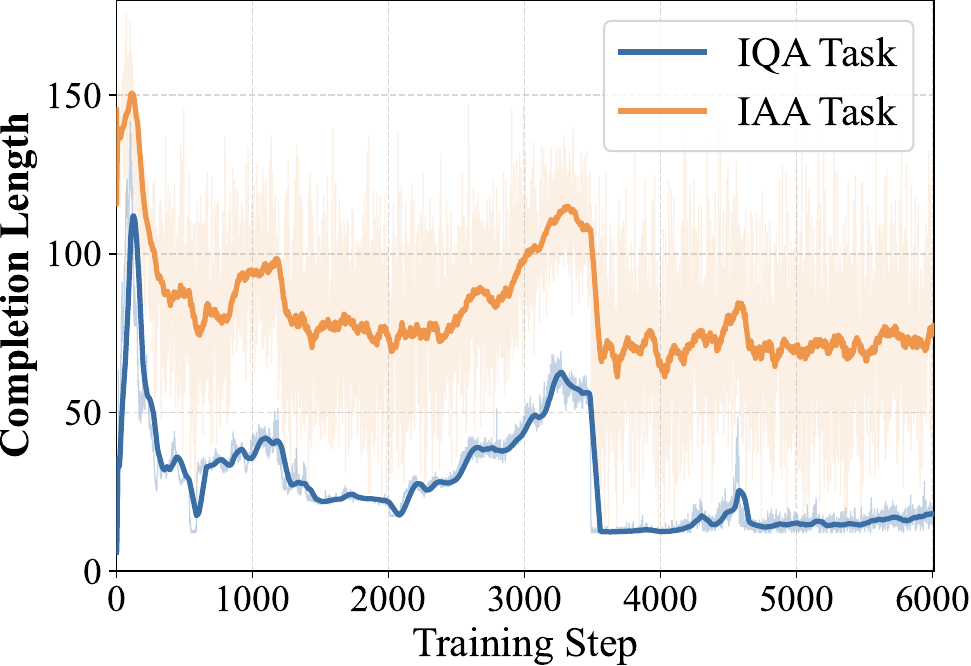}
        \caption{Divergent completion length}
        \label{fig:ReasoningCollapse}
    \end{subfigure}
    \caption{%
        \textbf{Motivation for task-conditioned post-training.}
        (a)~Adding a CoT instruction yields asymmetric effects: 
        it consistently helps IAA but hurts IQA, demonstrating 
        that a uniform reasoning strategy is suboptimal. 
        (b)~Under shared-reward RFT, IQA converges to short, 
        stable completions while IAA produces long, variable 
        ones, revealing that the two tasks inhabit different 
        response and optimization regimes.%
    }
    \label{fig:main_experimental_results}
\end{figure}

We verify both mismatches through targeted probes. In 
zero-shot evaluation on KonIQ-10k~\cite{KonIQ} and 
ArtiMuse~\cite{Artimuse}, simply prepending a chain-of-thought 
(CoT) instruction \emph{improves} IAA but \emph{degrades} IQA 
across four open-source MLLMs 
(Fig.~\ref{fig:CoTPerformanceOffset}), confirming that deliberative reasoning is beneficial for one task and harmful 
for the other. Separately, when we train a unified model under 
a shared reward, IQA completions collapse to short, stable 
outputs while IAA completions grow longer and more variable 
throughout training (Fig.~\ref{fig:ReasoningCollapse}). This 
divergence is not a side effect. It reveals that the two tasks 
occupy structurally different optimization landscapes that a 
single reward cannot simultaneously navigate well.

Motivated by these findings, we propose \underline{T}ask-\underline{A}ware \underline{T}hinking with \underline{A}symmetric \underline{R}ewards (\textbf{TATAR}), a unified MLLM framework that shares a single visual–language backbone while making post-training task-aware. TATAR has three components: (i) \textbf{Fast–slow reasoning construction}, where we build task-specific rationales (\texttt{QACoT-Score} dataset)—concise, distortion-focused explanations for IQA (via score-conditioned reverse inference) and longer, attribute-based aesthetic rationales for IAA (via structured summarization of expert annotations); (ii) \textbf{Two-stage reasoning-mode learning}, where a format-focused supervised 
fine-tuning (SFT) stage stabilizes task-conditioned outputs and reasoning length, followed by GRPO~\cite{shao2024deepseekmath} to refine scoring via group sampling and KL-regularized advantage optimization;  and (iii) \textbf{Asymmetric reward design}, using a Gaussian score reward for IQA and a Thurstone-style completion-level ranking reward for IAA to better match their supervision characteristics and improve training stability. Critically, the asymmetry is not an ad hoc design choice. It is the direct consequence of the two mismatches diagnosed above, and our ablations confirm that swapping rewards degrades both tasks.

Extensive experiments on eight benchmarks under both in-domain and 
cross-domain settings show that TATAR consistently outperforms prior 
unified baselines, remains competitive with task-specific  
models, and yields more stable training dynamics for aesthetics. Our results support a broader principle that effective unified perceptual 
scoring requires not a one-size-fits-all objective, but 
\emph{shared representation with task-conditioned adaptation}.

In summary, we make the following three contributions:
\begin{itemize}
    \item We identify two mismatches in unified IQA--IAA: a \emph{reasoning mismatch}, where the two tasks favor different response regimes, and an \emph{optimization mismatch}, where a shared point-wise objective is insufficient for aesthetic 
    judgment.

    \item We propose \textbf{TATAR}, a task-conditioned unified framework combining fast--slow reasoning construction, two-stage 
    SFT and GRPO training, and asymmetric Gaussian score and Thurstone ranking rewards to align IQA and IAA. 

    \item We conduct extensive experiments across $8$ benchmarks, demonstrating that task-conditioned post-training is an effective and principled paradigm for unified IQA--IAA.
\end{itemize}
\section{Related Work}

\subsection{Unified Image Quality and Aesthetic Assessment}

IQA~\cite{RALI,FRIQA,VQ_R1,DeQA_Doc,DeQA_score,EvoQuality,MPD} and 
IAA~\cite{photographer,AMHP,SRIAA,RPCD,Charm,Humanaesexpert} have 
traditionally been treated as separate problems, with task-specific 
architectures and objectives tailored to either distortion perception 
or aesthetic preference. IQA focuses on perceptual fidelity under 
degradations such as blur, noise, and compression~\cite{KonIQ}, while 
IAA evaluates subjective visual appeal shaped by composition, 
lighting, and semantics~\cite{IAAsurvey,Artimuse}.
The rise of MLLMs~\cite{Internvl3,Internvl35,llava15,Qwen25vl,Qwen3vl} 
has encouraged unified image scoring, where a single model handles 
both tasks. Existing approaches~\cite{Nexttoken,Qalign,UniPercept} 
share not only the backbone but also the training strategy, typically 
formulating both tasks as scalar score prediction or optimizing them 
under a common post-training objective. While these methods 
demonstrate that unified modeling is feasible, they treat IQA and IAA 
as homogeneous scoring problems and overlook their fundamentally 
different reasoning behaviors and supervision structures. Our work 
addresses this gap by sharing the backbone while conditioning 
post-training on the nature of each task.

\subsection{Post-training and Reward Design for MLLMs}

Post-training has become central to aligning 
MLLMs~\cite{DRL,ouyang2022training,silver2017mastering,shao2024deepseekmath,rafailov2023direct}. 
SFT, reinforcement learning, and preference-based optimization have 
all been shown to substantially improve reasoning and downstream 
performance in multimodal 
settings~\cite{MRN,GAPO,ShorterBetter,ToN,ARM,Perceptionr1,ReasonRFT}. 
A standard practice is a two-stage pipeline where SFT first 
establishes the desired response format, followed by reinforcement 
fine-tuning to improve task 
performance~\cite{MMUPT,SRPO,UnifiedMCOT}.
Reward design is a critical factor in this pipeline. Prior work has 
explored score-based rewards~\cite{Qinsight,ShorterBetter,GAPO,MRN}, 
preference optimization~\cite{UnifiedMCOT}, ranking-based 
objectives~\cite{VQ_R1}, and rule-based 
verifiers~\cite{Perceptionr1,ToN,SRPO}, collectively showing that 
reward quality often matters as much as the choice of optimization 
algorithm. However, unified scoring methods typically apply a single 
reward across both IQA and IAA~\cite{Qinsight,UniPercept}, ignoring 
their different supervision geometries. Our work instead applies 
asymmetric rewards: a score-oriented objective for IQA and a 
ranking-oriented objective for IAA, each matched to the task's 
underlying structure.

\section{Method}

\textbf{Motivation.} We propose TATAR, a unified framework for IQA 
and IAA built on a simple principle: \emph{the representation can be 
shared, but the post-training should be task-conditioned}. As 
discussed in Sec.~\ref{sec:intro}, IQA and IAA differ along two 
aspects. First, they favor different \emph{reasoning regimes}: IQA is 
distortion-centric and benefits from concise judgments, whereas IAA 
often requires more deliberative aggregation of semantic and 
compositional cues. Second, they differ in \emph{reward geometry}: 
IQA is naturally supervised by absolute quality scores, while IAA is 
better captured by the relative preference structure implied by 
aesthetic scores. TATAR addresses both mismatches within one MLLM 
policy.

\begin{figure}[t]
    \centering
    \includegraphics[width=\columnwidth]{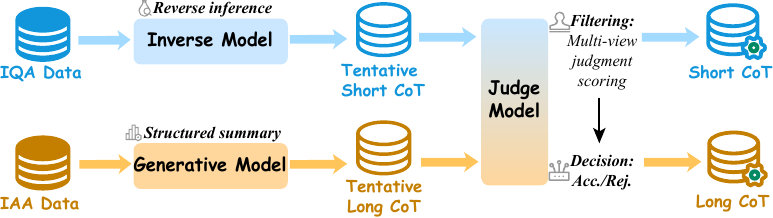}
    \caption{Task-conditioned reasoning construction. IQA rationales 
    are synthesized by score-conditioned reverse inference from 
    image--score pairs, while IAA rationales are obtained by 
    summarizing structured aesthetic annotations. A unified judge then 
    filters low-quality generations, producing the final fast--slow 
    reasoning corpus used in Stage~1 SFT.}
    \label{fig:data_pipeline}
\end{figure}

\begin{figure*}[t]
    \centering
    \includegraphics[width=0.95\textwidth]{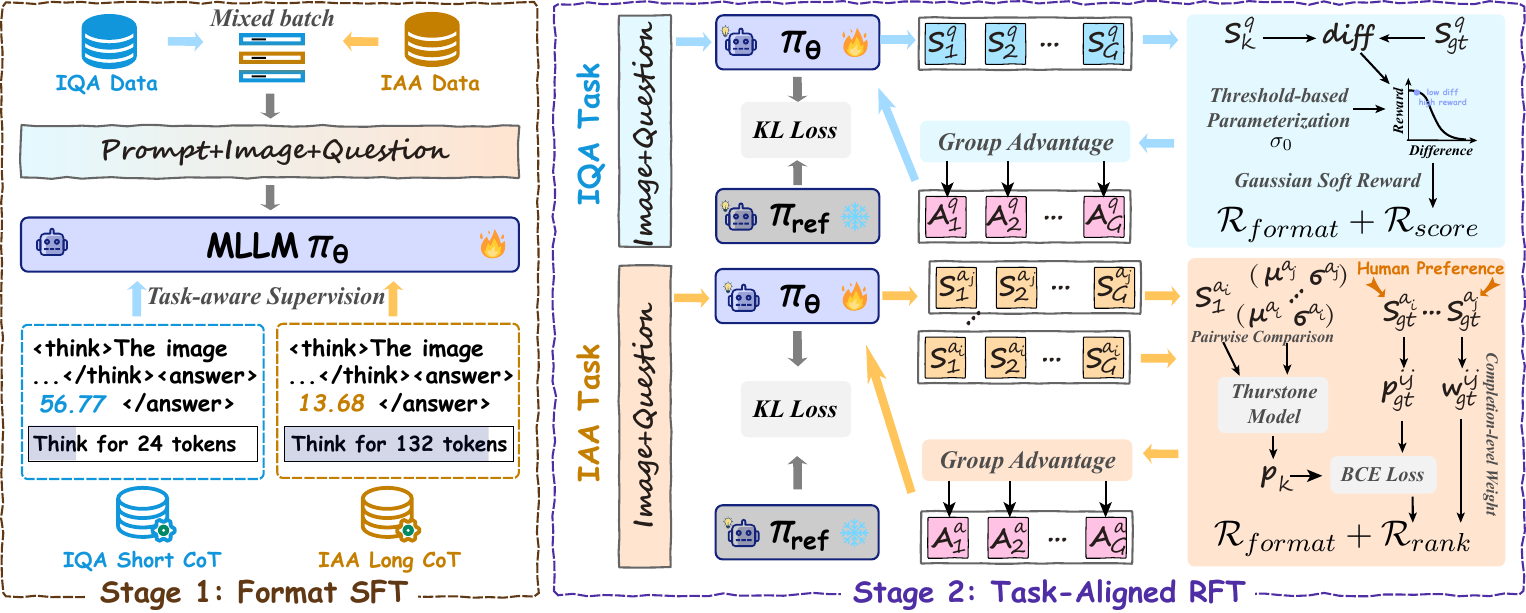}
    \caption{\textbf{Overview of TATAR.} TATAR is a unified framework 
    for IQA and IAA with task-conditioned reasoning and reward design. 
    In Stage~1, format-oriented SFT aligns the model to a shared 
    \texttt{<think>/<answer>} output schema while teaching distinct 
    response modes, with concise reasoning for IQA and more 
    deliberative reasoning for IAA. In Stage~2, task-conditioned GRPO 
    further refines the model with a shared optimization procedure but 
    asymmetric rewards, where IQA is optimized with a Gaussian score 
    reward $\mathcal{R}_{\mathrm{fmt}}+\mathcal{R}_{\mathrm{score}}$ 
    and IAA is optimized with a ranking reward 
    $\mathcal{R}_{\mathrm{fmt}}+\mathcal{R}_{\mathrm{rank}}$.}
    \label{fig:framework}
\end{figure*}

\vspace{0.5em}\noindent\textbf{Overview.} As shown in 
Fig.~\ref{fig:data_pipeline}\&\ref{fig:framework}, TATAR consists of three components. 
\textbf{(1) Fast--slow reasoning construction} (\S\ref{sec:dataconstruct}) 
builds task-specific demonstrations (\texttt{QACoT-score} dataset), pairing IQA with short perceptual 
rationales and IAA with longer aesthetic rationales. \textbf{(2) 
Task-conditioned two-stage learning} (\S\ref{sec:twostage}) first uses 
SFT to teach the model \emph{how} each task should be answered, then 
applies RFT to improve \emph{what} score is predicted. \textbf{(3) 
Asymmetric reward design} (\S\ref{sec:dualreward}) uses score-oriented 
optimization for IQA and ranking-oriented optimization for IAA. 
Importantly, the synthesized rationales are used only to initialize 
task-specific response priors in Stage~1; the final model is optimized 
in Stage~2 by task-aligned rewards.

\subsection{Fast--Slow Reasoning Construction}
\label{sec:dataconstruct}

\begin{figure}[t]
    \centering
    \includegraphics[width=\columnwidth]{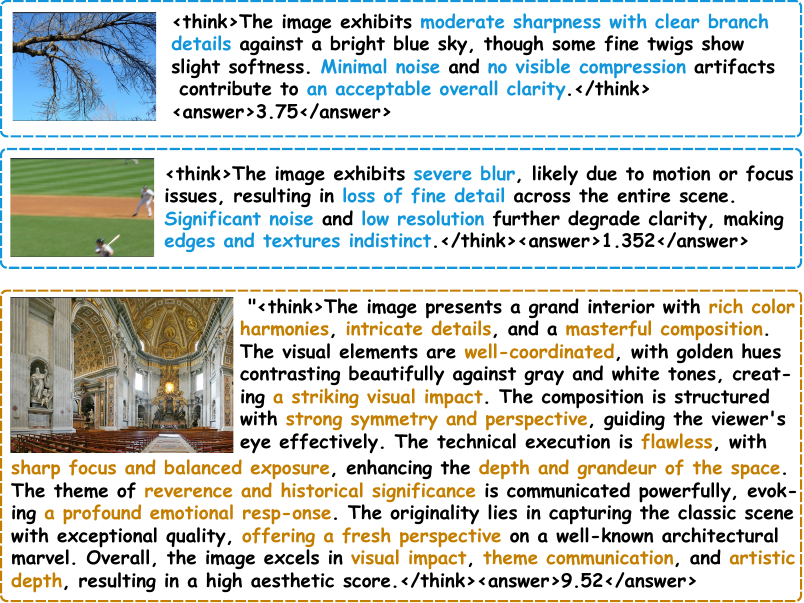}
    \caption{Two examples from the constructed reasoning corpus (\texttt{QACoT-score} dataset). IQA 
    (top) samples use concise, distortion-focused rationales, while 
    IAA (bottom) samples use longer and more integrative aesthetic 
    rationales.}
    \label{fig:cot_example}
\end{figure}

Existing IQA and IAA datasets provide reliable scalar scores but 
typically lack the task-specific reasoning traces needed to teach 
different response modes. We therefore construct a \emph{fast--slow} 
reasoning corpus--\texttt{QACoT-score} dataset--that reflects the distinct decision regimes of the 
two tasks. The construction is task-specific, but the output format is 
unified: each sample contains one \texttt{<think>} field and one 
\texttt{<answer>} field, where \texttt{<answer>} holds the final score.

\vspace{0.5em}\noindent\textbf{IQA: score-conditioned reverse 
inference.} Most IQA datasets contain only an image and its 
mean-opinion score (MOS). To obtain concise IQA rationales, we start 
from such image--score pairs and prompt a strong teacher MLLM to infer 
a short explanation grounded in visible distortions. Concretely, using 
KonIQ-10k~\cite{KonIQ} as the seed dataset, the teacher is given the 
image and its score and asked to produce a brief rationale focused on 
low-level quality factors such as blur, noise, compression artifacts, 
exposure, and detail loss. The rationale is intentionally short and 
distortion-oriented, reflecting the \emph{fast} decision regime of IQA.

\vspace{0.5em}\noindent\textbf{IAA: structured aesthetic 
summarization.} IAA requires a different form of supervision. We use 
ArtiMuse~\cite{Artimuse} as the seed dataset because it provides not 
only overall aesthetic scores but also rich attribute-level 
annotations including composition, technical execution, emotion, 
originality, and overall impression. Rather than inferring a rationale 
solely from the final score, we prompt a strong teacher model to 
summarize these structured attributes into a coherent long-form 
explanation and output the corresponding score. The goal is not to 
restate the annotations verbatim, but to organize them into a 
connected judgment process that reflects the more deliberative nature 
of aesthetic assessment.

\vspace{0.5em}\noindent\textbf{Unified judge-based filtering.} 
Synthetic rationales are inevitably noisy, so we apply a judge-based 
filtering step before training. A strong judge MLLM examines each 
generated sample and returns an accept or reject decision. The judge 
first applies shared hard constraints covering output format, score 
validity, and general reasoning cleanliness, and then performs 
task-specific checks. For IQA, it verifies whether the rationale is 
visually grounded and whether the identified distortions are plausible. 
For IAA, it checks whether the explanation covers relevant aesthetic 
factors, remains semantically coherent, and avoids unsupported 
conclusions. Only accepted samples are retained, yielding a clean 
task-aware reasoning corpus for Stage~1 training. Representative 
examples are shown in Fig.~\ref{fig:cot_example}.

\subsection{Task-Conditioned Two-Stage Learning}
\label{sec:twostage}

The constructed reasoning corpus specifies \emph{how} the model should 
respond to each task, but does not by itself guarantee optimal scoring 
behavior. We therefore train TATAR in two stages: a supervised stage 
that establishes task-specific response priors, followed by a 
reinforcement stage that improves prediction quality under 
task-aligned rewards.

\vspace{0.5em}\noindent\textbf{Stage 1: SFT for response-mode 
alignment.} In the first stage, we fine-tune a shared policy 
$\pi_{\theta}$ on mixed IQA and IAA batches under a unified 
\texttt{Prompt+Image+Question} interface. All outputs are normalized 
to the same schema,
\[
\texttt{<think>}\ \ldots\ \texttt{</think><answer>}\ \textit{score}\ 
\texttt{</answer>}.
\]
The key difference lies in the supervision inside \texttt{<think>}, 
where IQA demonstrations encourage short, perceptual rationales and 
IAA demonstrations encourage longer, more integrative reasoning. This 
stage teaches the model to adopt a task-appropriate response mode 
alongside predicting a score. In other words, Stage~1 establishes a behavioral prior for fast IQA reasoning and slow IAA reasoning, preventing the mode collapse observed under RL-only training 
(Fig.~\ref{fig:ReasoningCollapse}).

\vspace{0.5em}\noindent\textbf{Stage 2: GRPO for task-conditioned 
refinement.} Starting from the SFT-initialized policy, we further 
optimize the model using Group Relative Policy Optimization 
(GRPO)~\cite{shao2024deepseekmath}. For input $x = (I, q)$ consisting 
of an image and a task instruction, we sample a group of $G$ 
completions $\{y_k\}_{k=1}^{G}$ from the current policy and evaluate 
each with an external reward function $\mathcal{R}_k = 
\mathcal{R}(x, y_k)$. Rewards are normalized within the group to 
obtain relative advantages,
\begin{equation}
A_k = \frac{\mathcal{R}_k - \mathrm{mean}(\{\mathcal{R}_1, \ldots, 
\mathcal{R}_G\})}{\mathrm{std}(\{\mathcal{R}_1, \ldots, \mathcal{R}_G\}) 
+ \epsilon},
\label{eq:grpo_adv}
\end{equation}
where $\epsilon$ is a small constant for numerical stability. GRPO 
then updates the policy to increase the likelihood of high-advantage 
completions while regularizing toward a frozen reference model 
$\pi_{\mathrm{ref}}$ initialized from Stage~1,
\begin{align}
\mathcal{J}_{\mathrm{GRPO}}(\theta) = 
\mathbb{E}_{x \sim \mathcal{D},\, y_k \sim \pi_{\theta_{\mathrm{old}}}} 
\Bigg[ \frac{1}{G} \sum_{k=1}^{G} \min\Big(\rho_k(\theta) A_k,\, 
\nonumber \\
\mathrm{clip}(\rho_k(\theta), 1{-}\xi, 1{+}\xi) A_k \Big) 
- \beta\, D_{\mathrm{KL}}\!\left(\pi_{\theta} \| 
\pi_{\mathrm{ref}}\right) \Bigg],
\label{eq:grpo_obj}
\end{align}
where $\rho_k(\theta) = \pi_{\theta}(y_k \mid x) / 
\pi_{\theta_{\mathrm{old}}}(y_k \mid x)$. This formulation has two 
advantages in our setting. First, it directly optimizes end-task 
behavior without requiring a separate value model. Second, because 
Stage~1 has already established a valid response format, Stage~2 can 
focus on improving scoring quality rather than rediscovering output 
structure. The only component that differs across IQA and IAA in 
Stage~2 is the reward function, described next.

\subsection{Asymmetric Reward Design}
\label{sec:dualreward}

The optimizer is shared across tasks, but the reward should reflect 
the structure of each task. IQA is an absolute scoring problem where 
the goal is to predict a value close to the target MOS. IAA is less 
well modeled as strict point-wise regression, since multiple plausible 
aesthetic explanations may correspond to similar scores. We therefore 
use a shared format reward for both tasks and combine it with a 
task-specific reward matched to each supervision regime.

\vspace{0.5em}\noindent\textbf{Shared format reward.} For both IQA 
and IAA, we use a binary format reward 
$\mathcal{R}_{\mathrm{fmt}} \in \{0, 1\}$ that equals $1$ only when 
the completion contains exactly one pair of \texttt{<think>} and 
\texttt{<answer>} tags with a valid numeric score in the allowed 
range. This shared constraint discourages malformed outputs and 
ensures that task-specific rewards operate on comparable responses.

\vspace{0.5em}\noindent\textbf{IQA reward: Gaussian score shaping.} 
For IQA, the supervision is naturally score-based. Let $S_k^{q}$ be 
the predicted score of completion $k$ and $S_{\mathrm{gt}}^{q}$ the 
ground-truth MOS, with prediction error 
$\Delta_k^{q} = S_k^{q} - S_{\mathrm{gt}}^{q}$. Instead of a hard 
 threshold, we use a smooth Gaussian reward,
\begin{equation}
\mathcal{R}_{\mathrm{score}} = \exp\!\left( -\frac{(\Delta_k^{q})^2}
{2\,\sigma(\Delta_k^{q})^2} \right),
\label{eq:iqa_gauss}
\end{equation}
where the scale is mildly adaptive, 
$\sigma(\Delta_k^{q}) = \sigma_0(1 + \alpha|\Delta_k^{q}|/100)$. 
The base scale $\sigma_0$ is parameterized by an interpretable 
tolerance threshold $\tau$: if an absolute error of $\tau$ should receive 
reward $\eta$, then $\sigma_0 = \tau / \sqrt{-2\ln\eta}$. This 
design provides dense feedback near the target score while avoiding 
overly brittle optimization early in training. The final IQA reward is
\begin{equation}
\mathcal{R}_{\mathrm{IQA}} = 
\mathcal{R}_{\mathrm{fmt}} + \mathcal{R}_{\mathrm{score}}.
\label{eq:iqa_total}
\end{equation}

\vspace{0.5em}\noindent\textbf{IAA reward: score-induced relative 
ranking.} For IAA, we do not assume access to explicit pairwise 
preference annotations. Instead, we derive a soft relative preference 
from scalar aesthetic scores and optimize the model to preserve that 
relative structure. Consider IAA sample $i$ and completion $k$ 
predicting score $S_{k,i}^{a}$. Let $\mu_i^{a}$ and $(\sigma_i^{a})^2$ 
denote the mean and variance of predicted scores across the sampled 
completions of sample $i$. For another IAA sample $j$, the soft 
ground-truth preference is defined as
\begin{equation}
p_{\mathrm{gt}}^{ij} = \sigma\!\left( \frac{S_{\mathrm{gt},i}^{a} - 
S_{\mathrm{gt},j}^{a}}{\tau_a} \right),
\label{eq:iaa_gt}
\end{equation}
where $\sigma(\cdot)$ is the sigmoid function and $\tau_a$ controls 
the softness of the target. The predicted preference for completion 
$k$ of sample $i$ against sample $j$ follows a Thurstone-style 
comparison,
\begin{equation}
\hat{p}_{k}^{ij} = \sigma\!\left( \frac{S_{k,i}^{a} - \mu_j^{a}}
{\sqrt{(\sigma_i^{a})^2 + (\sigma_j^{a})^2 + \delta}} \right),
\label{eq:iaa_pred}
\end{equation}
where $\delta$ is a small constant for numerical stability. Since 
pairs with very different ground-truth scores are easy and less 
informative, we upweight ambiguous pairs via
\begin{equation}
w^{ij} = \exp\!\left( -\frac{|S_{\mathrm{gt},i}^{a} - 
S_{\mathrm{gt},j}^{a}|}{m} \right),
\label{eq:iaa_weight}
\end{equation}
where $m$ is a temperature parameter. The ranking loss for completion 
$k$ of sample $i$ is then
\begin{equation}
\mathcal{L}_{\mathrm{rank}}^{k,i} = \frac{\sum_{j \neq i} w^{ij}\, 
\mathrm{BCE}(p_{\mathrm{gt}}^{ij},\, \hat{p}_{k}^{ij})}
{\sum_{j \neq i} w^{ij}},
\label{eq:iaa_loss}
\end{equation}
which is converted to a reward by normalization, 
$\mathcal{R}_{\mathrm{rank}} = 1 - \mathcal{L}_{\mathrm{rank}}^{k,i} 
/ C_{\mathrm{rank}}$, where $C_{\mathrm{rank}}$ is a normalization constant used to scale the ranking reward. The total IAA reward is
\begin{equation}
\mathcal{R}_{\mathrm{IAA}} = 
\mathcal{R}_{\mathrm{fmt}} + \mathcal{R}_{\mathrm{rank}}.
\label{eq:iaa_total}
\end{equation}

\vspace{0.5em}\noindent\textbf{Discussion.} TATAR can be viewed as a 
unified model with task-conditioned post-training. The backbone, 
prompt interface, and optimization algorithm are shared, but the model 
is taught to respond differently and rewarded differently depending on 
whether the task is IQA or IAA. This design directly targets the two mismatches identified in Sec.~\ref{sec:intro}: reasoning mode is 
aligned via fast--slow SFT, and optimization signal is aligned via asymmetric rewards. Ablation in Sec.~\ref{sec:ablation} confirms that each contributes independently.
\section{Experiments}

\subsection{Experimental Setup}

\noindent\textbf{Datasets and Metrics.}
We train on the KonIQ-10k training split ($\sim$7,000 images) and the 
ArtiMuse-10K training set ($8,998$ images), following~\cite{DeQA_score,AMHP}. 
After applying the CoT generation and judge-based filtering pipeline 
described in Sec.~\ref{sec:dataconstruct}, we retain $6,189$ IQA 
samples and $7,726$ IAA samples, forming the QACoT-Score dataset. For 
evaluation, we use KonIQ-10k~\cite{KonIQ} and 
ArtiMuse-10K~\cite{Artimuse} as in-domain benchmarks, and 
SPAQ~\cite{SPAQ}, KADID-10k~\cite{kadid}, and PIPAL~\cite{Pipal} for 
out-of-domain IQA, and AVA~\cite{ava}, TAD66K~\cite{TAD66K}, and 
Flickr-AES~\cite{FLICKR} for out-of-domain IAA. All MOS labels are 
linearly normalized to a unified $1$--$100$ scale. We report Spearman 
Rank-order Correlation Coefficient (SRCC) and Pearson Linear 
Correlation Coefficient (PLCC) throughout, following ~\cite{UniPercept,Qalign,DeQA_score}.
\begin{table*}[ht!]
\centering
\small
\setlength{\tabcolsep}{3pt}
\renewcommand{\arraystretch}{0.995}
\caption{Performance comparison on unified IAA and IQA scoring. 
Results are reported as SRCC/PLCC (\%). For specialized models, 
\textcolor{red}{red} columns denote in-domain and 
\textcolor{green!50!black}{green} columns denote out-of-domain 
evaluation. $\clubsuit$ marks results cited from 
UniPercept~\cite{UniPercept}, $\spadesuit$ denotes single-task 
evaluation, and * denotes models retrained under our protocol.}
\begin{tabular}{l|ccccc|ccccc}
\toprule
\multirow{2}{*}{\textbf{Models}} 
& \multicolumn{5}{c|}{\textbf{IAA task}} 
& \multicolumn{5}{c}{\textbf{IQA task}} \\
\cmidrule(lr){2-6} \cmidrule(lr){7-11}
& \textbf{ArtiMuse-10K} & \textbf{AVA} & \textbf{TAD66K} 
& \textbf{FLICKR-AES} & \textbf{Avg.}
& \textbf{KonIQ-10K} & \textbf{SPAQ} & \textbf{KADID} 
& \textbf{PIPAL} & \textbf{Avg.} \\
\midrule
\textbf{\textit{Proprietary Models}} 
& \multicolumn{10}{c}{\textcolor{blue}{\textit{Large-scale 
pretraining, cross-domain testing}}} \\
GPT-4o~\cite{Gpt4} 
& 33.3/27.6 & 50.9/48.5 & 27.8/28.2 & 60.5/59.7 & 43.1/41.0
& 69.5/74.4 & 87.4/88.1 & 67.7/64.6 & 32.5/34.9 & 64.3/65.5 \\
Llama-4~\cite{Llama4scout}
& 20.4/14.7 & 34.5/32.9 & 23.6/21.0 & 54.8/50.6 & 33.3/29.8
& 50.3/65.3 & -4.1/0.7 & -9.9/-0.4 & -0.7/2.3 & 8.9/17.0 \\
Gemini-2.5-Pro~\cite{Gemini}
& 18.7/3.5 & 24.8/10.0 & 14.3/3.7 & 35.7/20.6 & 23.4/9.5
& 58.2/31.6 & 8.7/21.2 & 43.6/27.4 & 22.5/-1.9 & 33.3/19.6 \\
Claude-4.5~\cite{Claude35}
& 4.1/2.7 & 0.3/1.3 & 4.0/4.7 & 3.7/4.9 & 3.0/3.4
& -3.7/-4.3 & 3.6/8.5 & 22.3/27.3 & -13.1/-8.8 & 2.3/5.7 \\
\midrule
\textbf{\textit{Open-Source Models}} 
& \multicolumn{10}{c}{\textcolor{blue}{\textit{Large-scale 
pretraining, cross-domain testing}}} \\
LLaVA-1.5-8B~\cite{llava15} 
& 27.4/21.2 & 38.1/37.8 & 21.3/22.4 & 58.6/54.1 & 36.4/33.9
& 63.9/74.4 & -/- & 50.5/53.4 & 41.7/40.7 & 52.0/56.2 \\
GLM-4.5~\cite{GLM} 
& 34.6/24.9 & 46.4/42.0 & 28.9/27.8 & 65.1/59.7 & 43.8/38.6
& 72.1/76.5 & -4.0/-3.8 & -14.2/-12.8 & 1.3/2.0 & 13.8/15.5 \\
InternVL3-78B~\cite{Internvl3} 
& 22.3/20.6 & 38.5/34.4 & 22.1/22.0 & 51.8/43.3 & 33.7/30.1
& 63.5/67.6 & 84.9/85.2 & 57.9/55.3 & 41.5/45.7 & 61.9/63.4 \\
InternVL3.5-38B~\cite{Internvl35} 
& 21.9/17.5 & 35.9/35.7 & 20.1/20.8 & 55.9/52.9 & 33.4/31.7
& 57.8/65.2 & 84.0/83.1 & 56.8/53.7 & 44.8/45.7 & 60.8/61.9 \\
QwenVL-2.5-72B~\cite{Qwen25vl} 
& 23.3/19.7 & 40.8/38.7 & 23.2/23.5 & 62.6/58.9 & 37.5/35.2
& 76.2/82.0 & -/- & 60.6/57.0 & 38.1/40.7 & 58.3/59.9 \\
QwenVL-3-32B~\cite{Qwen3vl} 
& 22.7/13.0 & 35.3/19.8 & 20.0/9.5 & 57.2/41.3 & 33.8/20.9
& 79.6/83.8 & 69.0/65.7 & 67.3/68.2 & 41.4/40.2 & 64.3/64.4 \\
\midrule
\textbf{\textit{Specialized Models}}
& \textcolor{red}{\textit{In-domain}}
& \multicolumn{3}{c}{\textcolor{green!50!black}{\textit{Cross-domain}}}
& \textbf{Avg.}
& \textcolor{red}{\textit{In-domain}}
& \multicolumn{3}{c}{\textcolor{green!50!black}{\textit{Cross-domain}}}
& \textbf{Avg.} \\
ArtiMuse$\spadesuit$~\cite{Artimuse} 
& 61.4/62.7 & 39.7/38.5 & 23.0/23.2 & 34.9/33.4 & 39.8/39.5
& -/- & -/- & -/- & -/- & -/- \\
DeQA$\spadesuit$~\cite{DeQA_score} 
& -/- & -/- & -/- & -/- & -/-
& 95.3/94.1 & 89.5/89.6 & 69.4/68.7 & 47.2/47.8 & 75.3/75.0 \\
Q-Align$\clubsuit$~\cite{Qalign} 
& 55.1/57.3 & 39.8/38.6 & 19.4/19.7 & 13.7/12.3 & 32.0/32.0
& 94.1/94.0 & 88.6/88.7 & 67.4/68.4 & 40.3/41.9 & 72.6/73.3 \\
Q-Insight$\spadesuit$~\cite{Qinsight} 
& -/- & -/- & -/- & -/- & -/-
& 93.3/91.6 & 90.7/90.5 & 74.2/73.6 & 48.6/47.4 & 76.7/75.8 \\
Q-Insight$\clubsuit$~\cite{Qinsight} 
& 22.8/17.5 & 40.5/37.6 & 21.2/21.7 & 61.7/53.7 & 36.6/32.6
& 73.3/75.0 & 80.0/93.8 & 58.0/54.8 & 36.9/36.8 & 62.1/65.1 \\
\midrule
UniPercept*~\cite{UniPercept} 
& 43.2/45.1 & \textbf{53.5/50.9} & 30.6/31.2 & 52.1/53.4 & 44.8/45.1 
& 90.1/92.2 & 83.0/82.6 & 68.2/68.5 & 45.9/46.8 & 71.8/72.5 \\
\rowcolor{blue!15}
\textbf{TATAR (Ours)}
& \textbf{58.6/59.5} & 51.8/52.4 & \textbf{33.4/34.9} 
& \textbf{60.4/63.3} & \textbf{51.0/52.5} 
& \textbf{94.1/94.2} & \textbf{89.7/89.4} & \textbf{73.1/72.7} 
& \textbf{50.1/49.8} & \textbf{76.7/76.5} \\
\bottomrule
\end{tabular}
\label{tab:main_vr_results}
\end{table*}

\noindent\textbf{Evaluated Models.}
We compare against $19$ models across four categories. \emph{(1) 
Proprietary MLLMs:} GPT-4o~\cite{Gpt4}, 
Llama-4-Scout~\cite{Llama4scout}, Gemini-2.5-Pro~\cite{Gemini}, 
Claude-Sonnet-4.5~\cite{Claude35}, and 
Claude-Sonnet-4.5-Think~\cite{Claude35}. \emph{(2) Open-source 
MLLMs:} InternVL3 and InternVL3.5 series~\cite{Internvl3,Internvl35}, 
QwenVL-2.5 and QwenVL-3 series~\cite{Qwen25vl,Qwen3vl}, 
GLM-4.5V~\cite{GLM}, and LLaVA-1.5~\cite{llava15}. \emph{(3) 
Specialized models:} Q-Align~\cite{Qalign}, ArtiMuse~\cite{Artimuse}, 
DeQA~\cite{DeQA_score}, and Q-Insight~\cite{Qinsight}. \emph{(4) 
Unified scoring:} UniPercept~\cite{UniPercept}, retrained under the 
same data and protocol as ours for a fair comparison.

\noindent\textbf{Implementation Details.}
We adopt Qwen-2.5-VL-7B-Instruct~\cite{Qwen25vl} as the base model for all experiments.
For GRPO, we set the number of sampled generations per input to $N=8$, the KL penalty weight to $\beta=10^{-3}$.
We use AdamW with an initial learning rate of $10^{-6}$, linearly decayed to $10^{-9}$. 
Stage 1 (Format SFT) is trained for $1$ epoch, followed by Stage 2 RFT for $2$--$3$ epochs, with a total batch size of $8{\times}2{\times}8$ on $8$ NVIDIA A800 GPUs.
For heterogeneous image resolutions across IQA/IAA datasets, we cap the maximum number of image pixels to $1{,}572{,}864$ pixels. 
For the Gaussian score reward, we set $\alpha_{\text{IQA}}=0.8$, $\alpha_{\text{IAA}}=2.0$ (If used in Ablation), the target level $\eta=0.5$, and the difference threshold $\tau$ to 8.75.
For the ranking reward, the soft preference temperature $\tau$ is set to $0.08$, and the hard-pair weighting temperature $m$ is set to $0.12$.

\begin{table*}[ht!]
\centering
\small
\setlength{\tabcolsep}{4pt}
\renewcommand{\arraystretch}{1}
\caption{Ablation study on training paradigm and reward design. 
Results are reported as SRCC/PLCC (\%). \textbf{Bold} and 
\underline{underlined} values indicate first and second place. 
Task-cond.\ means Gaussian for IQA and Rank for IAA.}
\label{tab:ablation}
\begin{tabular}{ccc|ccccc|ccccc}
\toprule
\multicolumn{3}{c|}{\textbf{Settings}} 
& \multicolumn{5}{c|}{\textbf{IAA task}} 
& \multicolumn{5}{c}{\textbf{IQA task}} \\
\cmidrule(lr){1-3}\cmidrule(lr){4-8}\cmidrule(lr){9-13}
\textbf{SFT} & \textbf{RL} & \textbf{Reward}
& \textbf{ArtiMuse} & \textbf{AVA} & \textbf{TAD66K} 
& \textbf{FLICKR} & \textbf{Avg.}
& \textbf{KonIQ} & \textbf{SPAQ} & \textbf{KADID} 
& \textbf{PIPAL} & \textbf{Avg.} \\
\midrule
\checkmark & & 
& 45.9/45.1 & 30.3/34.2 & 18.9/19.7 & 36.3/37.8 & 32.9/34.2
& 89.2/91.8 & 81.5/82.3 & 70.9/69.0 & 40.2/41.2 & 70.5/71.1 \\
\midrule
& \checkmark & Hard-only
& 40.0/43.1 & \textbf{56.1/55.1} & 30.5/31.3 & 53.4/53.4 & 45.0/45.7
& 90.1/92.2 & 82.1/81.7 & 70.3/68.5 & 41.3/41.5 & 70.9/71.0 \\
& \checkmark & Gauss.-only
& 40.8/43.7 & 52.1/52.3 & 31.2/31.9 & 54.0/54.1 & 44.5/45.5
& 90.8/92.6 & 82.6/82.1 & 69.9/68.2 & 42.0/42.1 & 71.3/71.2 \\
& \checkmark & Rank-only
& 41.3/44.1 & \underline{55.8/51.4} & 30.9/31.6 & 54.6/54.8 & 45.6/45.5
& 91.2/92.9 & 82.3/81.9 & 70.8/69.1 & 42.6/42.8 & 71.7/71.7 \\
& \checkmark & Task-cond.
& 41.7/44.6 & 55.6/51.3 & 31.3/32.0 & 55.0/55.2 & 46.0/45.9
& 91.7/93.2 & 82.8/82.3 & 71.3/69.6 & 43.0/43.2 & 72.2/72.1 \\

\midrule
\checkmark & \checkmark & Hard-only
& 55.5/53.8 & 46.6/43.5 & 30.0/29.8 & 56.6/52.4 & 47.2/44.9
& 90.1/91.4 & 84.0/85.6 & 71.7/66.6 & 43.1/43.5 & 72.2/71.8 \\
\checkmark & \checkmark & Gauss.-only
& 56.6/57.5 & 47.9/49.1 & 30.2/32.7 
& 58.4/59.7 & 49.7/50.7
& 92.5/93.6 & 88.2/87.9 & \underline{73.2/70.0} 
& 45.9/47.2 & 75.0/74.7 \\
\checkmark & \checkmark & Rank-only
& \underline{57.8/58.6} & \underline{50.3/50.9} & \underline{31.8/33.9} & \underline{59.2/61.0}
& \underline{50.3/51.6}
& \underline{93.3/93.9} & \underline{88.9/88.6} & 73.0/71.5 & \underline{47.8/48.3}
& \underline{75.4/75.6} \\
\checkmark & \checkmark & Task-cond.
& \textbf{58.6/59.5} & 51.8/52.4 & \textbf{33.4/34.9} 
& \textbf{60.4/63.3} & \textbf{51.0/52.5}
& \textbf{94.1/94.2} & \textbf{89.7/89.4} & \textbf{73.1/72.7} 
& \textbf{50.1/49.8} & \textbf{76.7/76.5} \\
\bottomrule
\end{tabular}
\end{table*}

\subsection{Main Results}

Tab.~\ref{tab:main_vr_results} reports SRCC/PLCC on all eight 
benchmarks. TATAR achieves the best average performance on both tasks, 
with an IAA average of $51.0$/$52.5$ and an IQA average of 
$76.7$/$76.5$. Compared with the unified baseline UniPercept, TATAR 
improves IAA by $+6.2$/$+7.4$ points and IQA by $+4.9$/$+4.0$ points, 
demonstrating that task-conditioned post-training consistently 
outperforms a shared optimization recipe across both tasks and all 
evaluation settings.

\noindent\textbf{Comparison with proprietary models.}
Among proprietary models, GPT-4o achieves the most competitive 
results, with an IAA average of $43.1$/$41.0$ and an IQA average of 
$64.3$/$65.5$, but still falls short of TATAR by $7.9$/$11.5$ points 
on IAA and $12.4$/$11.0$ points on IQA, despite having no training on 
the evaluation domains. Gemini-2.5-Pro, Llama-4-Scout, and 
Claude-Sonnet-4.5 show substantially degraded performance, with 
several benchmarks yielding near-zero or negative SRCC. We attribute 
this to score-range misalignment: these models tend to output scores 
concentrated in a narrow band that does not align with the annotation 
scale, leading to low correlation regardless of 
underlying capability. This is a prompt-induced failure mode rather 
than a fundamental limitation, but it highlights the sensitivity of zero-shot scoring to output calibration.

\noindent\textbf{Comparison with open-source models.}
Open-source MLLMs show a clear task asymmetry: models such as 
QwenVL-3-32B achieve competitive IQA performance ($79.6$/$83.8$ 
in-domain) but weak IAA performance ($33.8$/$20.9$ average), 
consistent with the observation that aesthetic judgment requires 
higher-level semantic integration that general-purpose pretraining 
does not reliably provide. GLM-4.5 shows the opposite pattern on IQA: 
strong in-domain performance ($72.1$/$76.5$) but severe degradation 
on out-of-domain benchmarks including negative SRCC on KADID 
($-14.2$/$-12.8$), suggesting poor cross-domain generalization. TATAR 
improves over all open-source models on both tasks and generalizes 
substantially better across domains, with a $+7.2$/$+8.3$ IQA average 
gain over the strongest open-source model QwenVL-3-32B.

\noindent\textbf{Comparison with specialized models.}
On IQA, TATAR matches the in-domain performance of the best 
specialized model DeQA ($94.1$/$94.2$ vs. $95.3$/$94.1$) while 
simultaneously handling IAA, which DeQA cannot. On out-of-domain IQA, 
TATAR outperforms DeQA on SPAQ ($89.7$/$89.4$ vs. $89.5$/$89.6$), 
KADID ($73.1$/$72.7$ vs. $69.4$/$68.7$), and PIPAL 
($50.1$/$49.8$ vs. $47.2$/$47.8$). On IAA, TATAR substantially 
surpasses the specialized ArtiMuse model on cross-domain benchmarks 
($51.0$/$52.5$ unified average vs. $39.8$/$39.5$), with particularly 
strong gains on Flickr-AES ($60.4$/$63.3$ vs. $34.9$/$33.4$) and 
ArtiMuse in-domain ($58.6$/$59.5$ vs. $61.4$/$62.7$, competitive 
despite being a unified model). These results confirm that 
task-conditioned unification does not sacrifice task-specific accuracy 
for breadth.

\subsection{Ablation Studies}
\label{sec:ablation}

\begin{figure*}[t]
    \centering
    \begin{subfigure}[b]{0.24\linewidth}
        \centering
        \includegraphics[width=\linewidth]{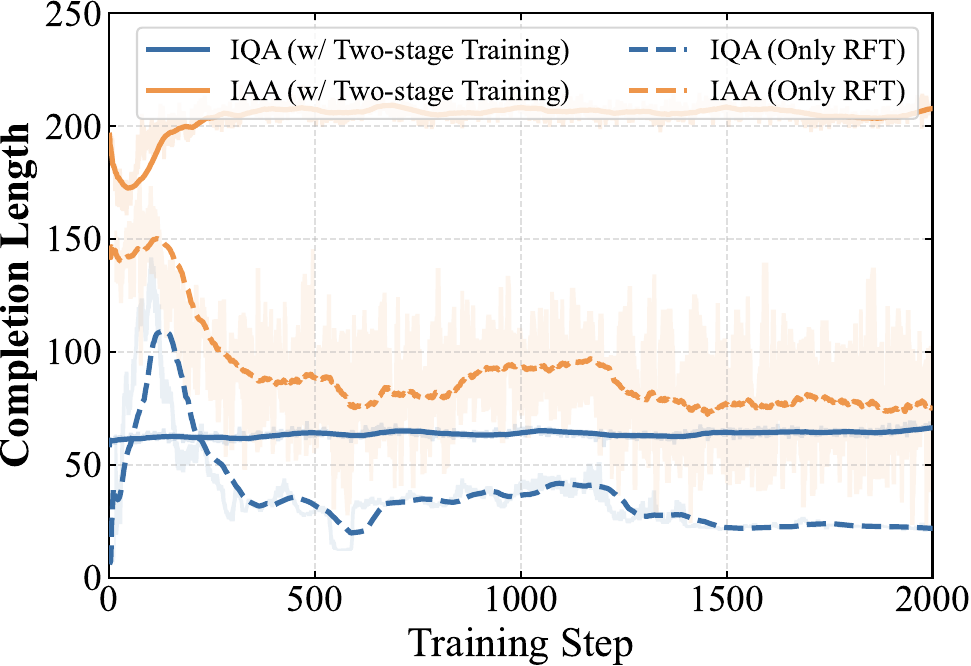}
        \caption{Effect of two-stage training on completion length.}
        \label{fig:Completion_Comparison}
    \end{subfigure}
    \hfill
    \begin{subfigure}[b]{0.24\linewidth}
        \centering
        \includegraphics[width=\linewidth]{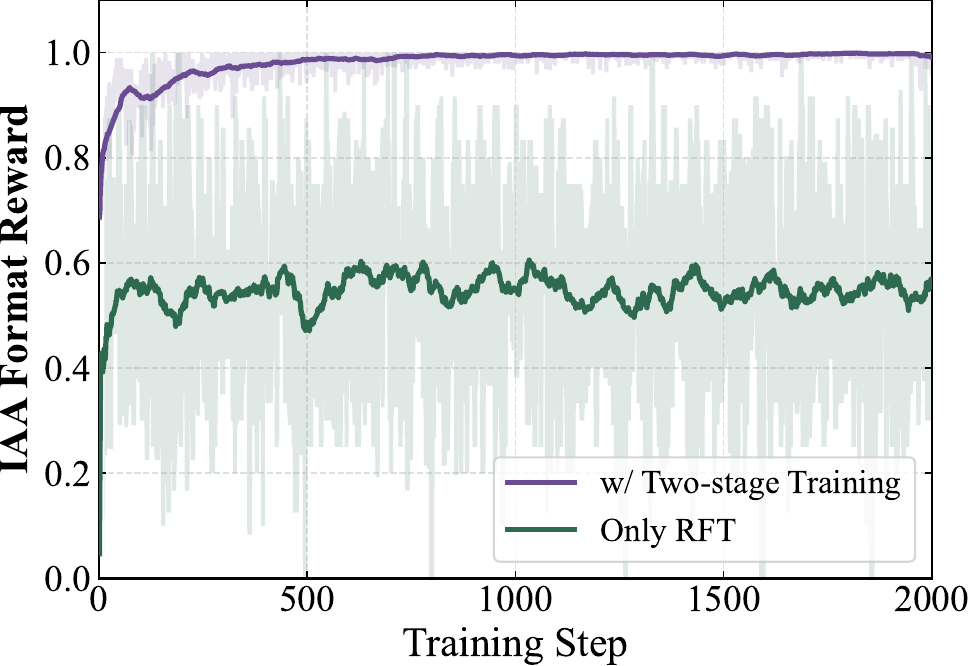}
        \caption{Effect of two-stage training on IAA format reward.}
        \label{fig:IAA_Format_Reward_Comparison}
    \end{subfigure}
    \hfill
    \begin{subfigure}[b]{0.24\linewidth}
        \centering
        \includegraphics[width=\linewidth]{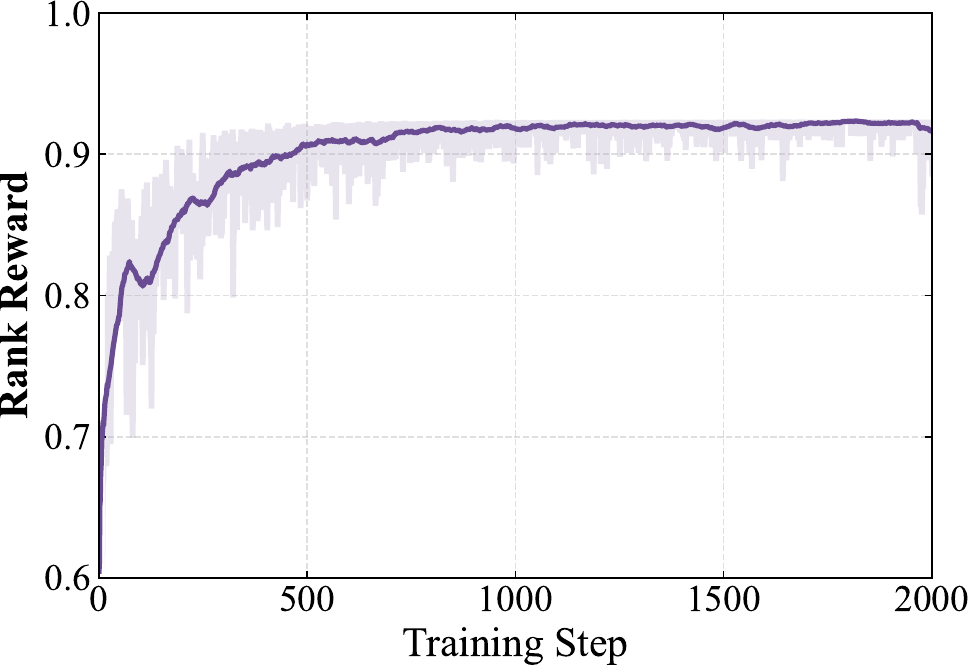}
        \caption{Rank reward trend during training.}
        \label{fig:Rank_Reward_Trend_Final}
    \end{subfigure}
    \hfill
    \begin{subfigure}[b]{0.24\linewidth}
        \centering
        \includegraphics[width=\linewidth]{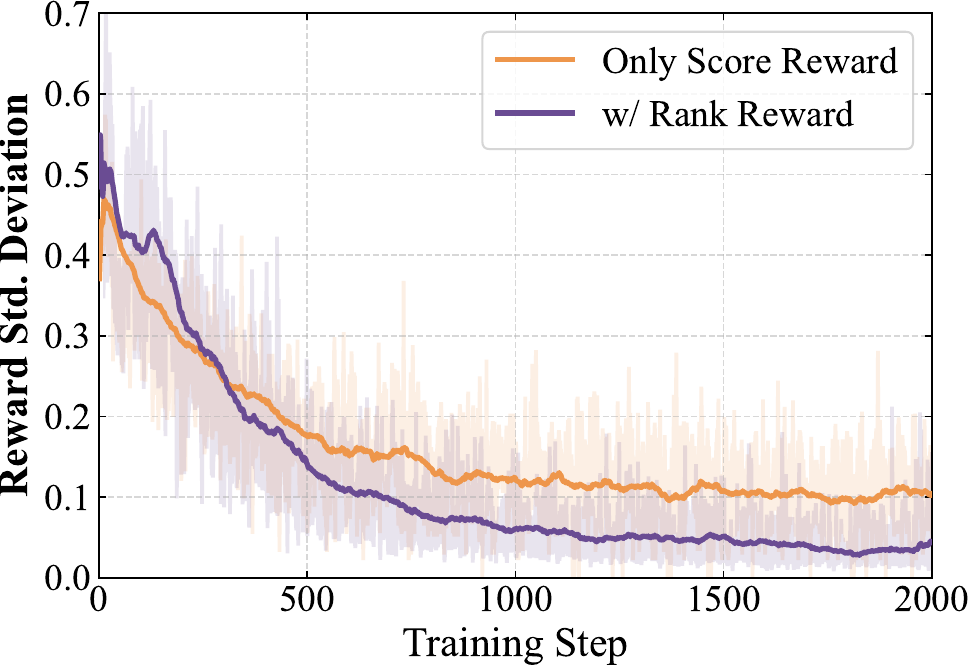}
        \caption{Effect of rank reward on reward standard deviation.}
        \label{fig:RewardStdCompare}
    \end{subfigure}
    \caption{Training dynamics analysis covering completion behavior, 
    reward trends, and optimization stability.}
    \label{fig:training_analysis}
\end{figure*}

\noindent\textbf{Effect of training paradigm.}
We compare three configurations: SFT only, RL only, and the full 
two-stage SFT+RL. SFT alone yields a weak IAA average of $32.9$/$34.2$, 
far below any RL-based configuration. This is expected: supervised 
imitation on the reasoning corpus teaches response format but cannot 
directly align the scoring policy with the target metrics. RL-only 
training recovers substantially, reaching IAA averages of 
$44.5$--$46.0$/$45.5$--$45.9$ across reward variants, but at the cost 
of output instability: completion lengths collapse and format rewards 
stay low (Fig.~\ref{fig:IAA_Format_Reward_Comparison}), indicating 
the model degenerates toward short, reward-seeking responses without 
SFT's structural prior. On IQA, RL-only achieves $70.9$--$72.2$/$71.0$--$72.1$, 
which is already competitive but still consistently below the best 
two-stage configurations. Combining SFT with RL yields the strongest 
overall performance, with the full configuration reaching $51.0$/$52.5$ 
on IAA and $76.7$/$76.5$ on IQA. Compared with the best RL-only variant, 
this corresponds to gains of $+5.0$/$+6.6$ on IAA and $+4.5$/$+4.4$ on IQA. 
These results confirm that SFT and RL are complementary: SFT establishes 
a stable behavioral prior, which RL then refines toward task-aligned scoring.

\noindent\textbf{Effect of reward design.}
We compare four reward configurations: Hard-only, Gauss.-only, 
Rank-only, and the Task-cond. Within RL-only training, 
replacing the hard threshold with the Gaussian reward brings a 
modest but consistent improvement on IQA ($71.3$/$71.2$ vs.\ $70.9$/$71.0$), 
while slightly reducing IAA performance ($44.5$/$45.5$ vs.\ $45.0$/$45.7$). 
Using Rank-only improves over Gauss.-only on both tasks, reaching 
$45.6$/$45.5$ on IAA and $71.7$/$71.7$ on IQA, suggesting that ranking 
supervision is already an effective signal even without explicit 
score regression. Using the task-conditioned reward (Task-cond.), 
which applies Gaussian score reward to IQA and ranking reward to IAA, 
further improves RL-only performance to $46.0$/$45.9$ on IAA and 
$72.2$/$72.1$ on IQA, indicating that the two reward forms are 
complementary when matched to the corresponding task.
Under the SFT+RL setting, the reward differences become more pronounced. 
Hard-only reaches $47.2$/$44.9$ on IAA and $72.2$/$71.8$ on IQA. 
Switching to Gauss.-only improves both tasks substantially, reaching 
$49.7$/$50.7$ on IAA and $75.0$/$74.7$ on IQA, corresponding to gains 
of $+2.5$/$+5.8$ and $+2.8$/$+2.9$, respectively. Rank-only further 
outperforms Gauss.-only, achieving $50.3$/$51.6$ on IAA and 
$75.4$/$75.6$ on IQA, showing that relative ranking becomes especially 
effective once the model is initialized by SFT. Finally, Task-cond. 
achieves the best overall results, reaching $51.0$/$52.5$ on IAA and 
$76.7$/$76.5$ on IQA, improving over Gauss.-only by $+1.3$/$+1.8$ on 
IAA and $+1.7$/$+1.8$ on IQA. This confirms that task-matched reward 
design provides the most effective optimization signal in the two-stage setting.

\begin{figure*}[t]
    \centering
    \includegraphics[width=\textwidth]{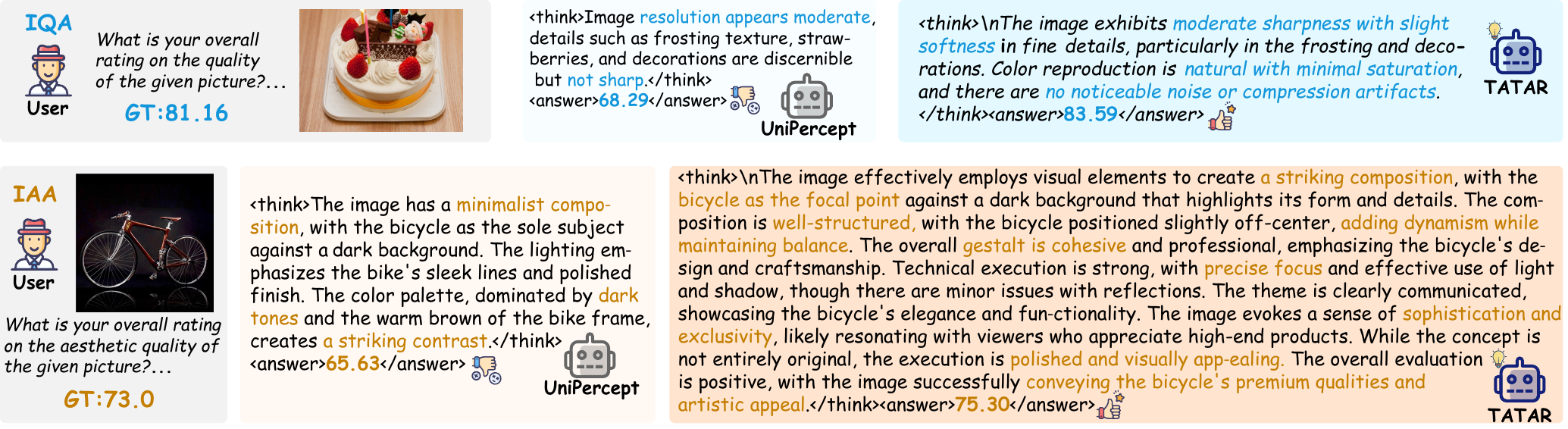}
    \caption{Case study comparing TATAR and UniPercept on 
    representative IQA (top) and IAA (bottom) examples. Ground-truth 
    scores and prediction errors are shown for each model.}
    \label{fig:casestudy}
\end{figure*}
\subsection{Training Dynamics Analysis}

\noindent\textbf{Effect of two-stage training.}
Fig.~\ref{fig:Completion_Comparison} shows that under RL-only 
training, IQA completion lengths collapse rapidly toward near-zero 
within the first $\sim$$500$ steps, while IAA completions fluctuate 
without converging. This collapse is the direct behavioral consequence 
of the reasoning mismatch: without a prior that distinguishes fast and 
slow response modes, RL uniformly minimizes completion cost. Two-stage 
training maintains substantially longer and more stable completions 
for both tasks throughout all $2{,}000$ training steps, with IAA 
completions remaining noticeably longer than IQA completions as 
intended. Fig.~\ref{fig:IAA_Format_Reward_Comparison} reinforces this 
picture: with two-stage training, the IAA format reward rises to near 
$1.0$ within the first $\sim$$200$ steps and stays stable, whereas 
RL-only training settles at a substantially lower plateau of around 
$0.5$--$0.6$ with high variance throughout. This shows that SFT 
initialization is necessary not only for preserving reasoning length, 
but for maintaining structural output validity during the RL phase.

\noindent\textbf{Effect of rank reward.}
Fig.~\ref{fig:Rank_Reward_Trend_Final} shows that the rank reward 
increases steadily from around $0.65$ at initialization to above $0.90$ 
at convergence, indicating that the model progressively learns 
stronger relative discrimination among candidate completions. The 
smooth and monotonic increase suggests that the Thurstone-style 
pairwise objective provides stable gradients throughout training, 
without the oscillation often seen in preference-based optimization. 
Fig.~\ref{fig:RewardStdCompare} shows that introducing the rank reward 
reduces the standard deviation of the overall reward by approximately 
$40$--$50\%$ compared with score-only training, particularly in the 
early and mid phases of RL. This reduction in reward variance directly 
translates to more stable policy updates, confirming that ranking 
supervision mitigates the reward oscillation that pure score regression 
introduces on subjective aesthetic data. Together, these dynamics 
support the core design choice: score-based supervision anchors 
absolute prediction quality while rank-based supervision stabilizes 
the relative preference structure that aesthetic assessment requires.
\subsection{Case Study}

Fig.~\ref{fig:casestudy} compares TATAR and UniPercept on 
representative IQA and IAA examples. On the IQA example, UniPercept 
produces a coarse judgment based on local discernibility, with a 
prediction error of $12.87$ points. TATAR explicitly identifies 
sharpness, fine-detail softness, color naturalness, and compression 
artifacts, reducing the error to $2.43$ points. This reflects the 
fast, distortion-focused reasoning mode instilled by Stage~1 SFT.
On the IAA example, the gap is more pronounced. UniPercept relies on 
three generic visual cues and underestimates the ground truth by 
$7.37$ points. TATAR produces a more integrative assessment covering 
composition balance, focal emphasis, craftsmanship, and thematic 
expression, reducing the error to $2.30$ points. This deliberative 
reasoning reflects the slow IAA mode elicited by our fast--slow 
construction and asymmetric ranking reward, and is consistent with 
TATAR's quantitative gains over UniPercept in 
Tab.~\ref{tab:main_vr_results}.

\section{Conclusion}
In this work, we revisited unified Image Quality Assessment (IQA) and Image Aesthetic Assessment (IAA) from the perspective of \emph{task-aware reasoning and optimization}. We identified a critical mismatch in prior unified training pipelines: IQA favors concise, distortion-grounded evidence, whereas IAA requires more deliberative, semantics-driven judgment and is poorly served by a uniform regression-style objective. To address this, we propose \textbf{TATAR}, a unified MLLM framework that explicitly separates task-specific \emph{thinking modes} and \emph{optimization signals}: (i) fast--slow CoT construction for IQA and IAA, (ii) two-stage training with Format SFT followed by GRPO-based RFT, and (iii) asymmetric rewards with Gaussian score shaping for IQA and Thurstone-style completion-level ranking for IAA. Overall, our results suggest that effective unified perceptual scoring requires \emph{task-aware reasoning control} and \emph{reward alignment} rather than a one-size-fits-all objective. In future work, we plan to extend TATAR to broader perceptual tasks (e.g., video quality and aesthetics), explore more faithful and scalable preference supervision, and study continual adaptation under real-world distribution shifts.


\bibliographystyle{ACM-Reference-Format}
\bibliography{yin}









\end{document}